\documentclass[letterpaper, 10 pt, conference]{ieeeconf}  
\IEEEoverridecommandlockouts                  
\usepackage[letterpaper, left=0.75in, right=0.75in, bottom=0.75in, top=1in]{geometry}
\usepackage{graphics} 
\usepackage{amsmath}  
\usepackage{amssymb}  
\usepackage[export]{adjustbox} 
\usepackage{bm}
\usepackage{balance}
\usepackage{subcaption} 
\usepackage{caption}
\usepackage{notoccite}
\usepackage{CJKutf8}
\usepackage{tikz}
\usepackage[compact]{titlesec}
\usepackage{algorithm}
\usepackage{algpseudocode}
\usepackage{stfloats} 
\usepackage{setspace} 

\makeatletter
 \let\NAT@parse\undefined
 \makeatother
\usepackage[pdftex,
        colorlinks=true,
        bookmarks=true,
        citecolor=red,
        linkcolor=blue,
        pagebackref=false,
        pdfstartview=Fit,
        breaklinks=true
            ]{hyperref}
\usepackage{cleveref}

\title{\LARGE\bf
Real-Time Polygonal Semantic Mapping for Humanoid Robot Stair Climbing
}
\author{Teng Bin$^{1,2}$, Jianming Yao$^{1,3}$, Tin Lun Lam$^{1,4}$,  Tianwei Zhang$^{1*}$
\thanks{This work was supported by the National Natural Science Foundation of China (Grant No. 62306185); 
the Guangdong Basic and Applied Basic Research Foundation (Grant No. 2024A1515012065), the Shenzhen Science and Technology Program (Grant No. JSGGKQTD20221101115656029 and KJZD20230923113801004)}
\thanks{$^{1}$The Shenzhen Institute of Artificial Intelligence and Robotics for Society, Shenzhen, China}
\thanks{$^{2}$College of Intelligent Systems Science and Engineering, Harbin Engineering University, Harbin, China}
\thanks{$^{3}$Electronic science and technology, Guangdong University of Technology, Guangzhou, China}
\thanks{$^{4}$The Chinese University of Hong Kong - Shenzhen, Shenzhen, China}
\thanks{* Corresponding Author: {zhangtianwei@cuhk.edu.cn}}
}

\begin{document}
\begin{CJK}{UTF8}{gbsn}
\maketitle
\thispagestyle{empty}
\pagestyle{empty}


\begin{abstract}

 We present a novel algorithm for real-time planar semantic mapping tailored for humanoid robots navigating complex terrains such as staircases. Our method is adaptable to any odometry input and leverages GPU-accelerated processes for planar extraction, enabling the rapid generation of globally consistent semantic maps. We utilize an anisotropic diffusion filter on depth images to effectively minimize noise from gradient jumps while preserving essential edge details, enhancing normal vector images' accuracy and smoothness. Both the anisotropic diffusion and the RANSAC-based plane extraction processes are optimized for parallel processing on GPUs, significantly enhancing computational efficiency. Our approach achieves real-time performance, processing single frames at rates exceeding $30~Hz$, which facilitates detailed plane extraction and map management swiftly and efficiently. Extensive testing underscores the algorithm’s capabilities in real-time scenarios and demonstrates its practical application in humanoid robot gait planning, significantly improving its ability to navigate dynamic environments.
\end{abstract}
\section{Introduction}\label{char1}

The capability to perceive the environment accurately and construct effective traversable semantic maps is crucial for humanoid robots navigating complex environments. In particular, for humanoid robots, securing stable footholds in sufficiently large planar areas is essential for safety and accuracy in movement planning. Due to their greater degrees of freedom, humanoid robots can undertake more complex tasks, such as jumping and climbing stairs, compared to traditional wheeled robots. However, these complex movements are based on the robot's ability to precisely and efficiently sense its surroundings and build reliable semantic maps. Hence, enhancing planar semantic mapping for humanoid robots is paramount and warrants further research and discussion.

The construction of reliable, traversable planar semantic maps for humanoid robots has long captured the attention of multiple disciplines including computer vision and SLAM (Simultaneous Localization and Mapping). In the domain of computer vision, numerous effective algorithms have been proposed to enhance performance metrics such as real-time responsiveness and Intersection over Union (IOU) by focusing on extracting planes from ordered structures like depth images, which generally perform better in real-time and accuracy than methods based on unordered point clouds \cite{4}. Additionally, the field of SLAM has utilized planar and curved spatial semantic information for robot localization and mapping \cite{3,82}, with several successful projects now open-sourced, such as elevation mapping with GPU \cite{18}.


\begin{figure}[tbp]
\centering
\includegraphics[width=\columnwidth]{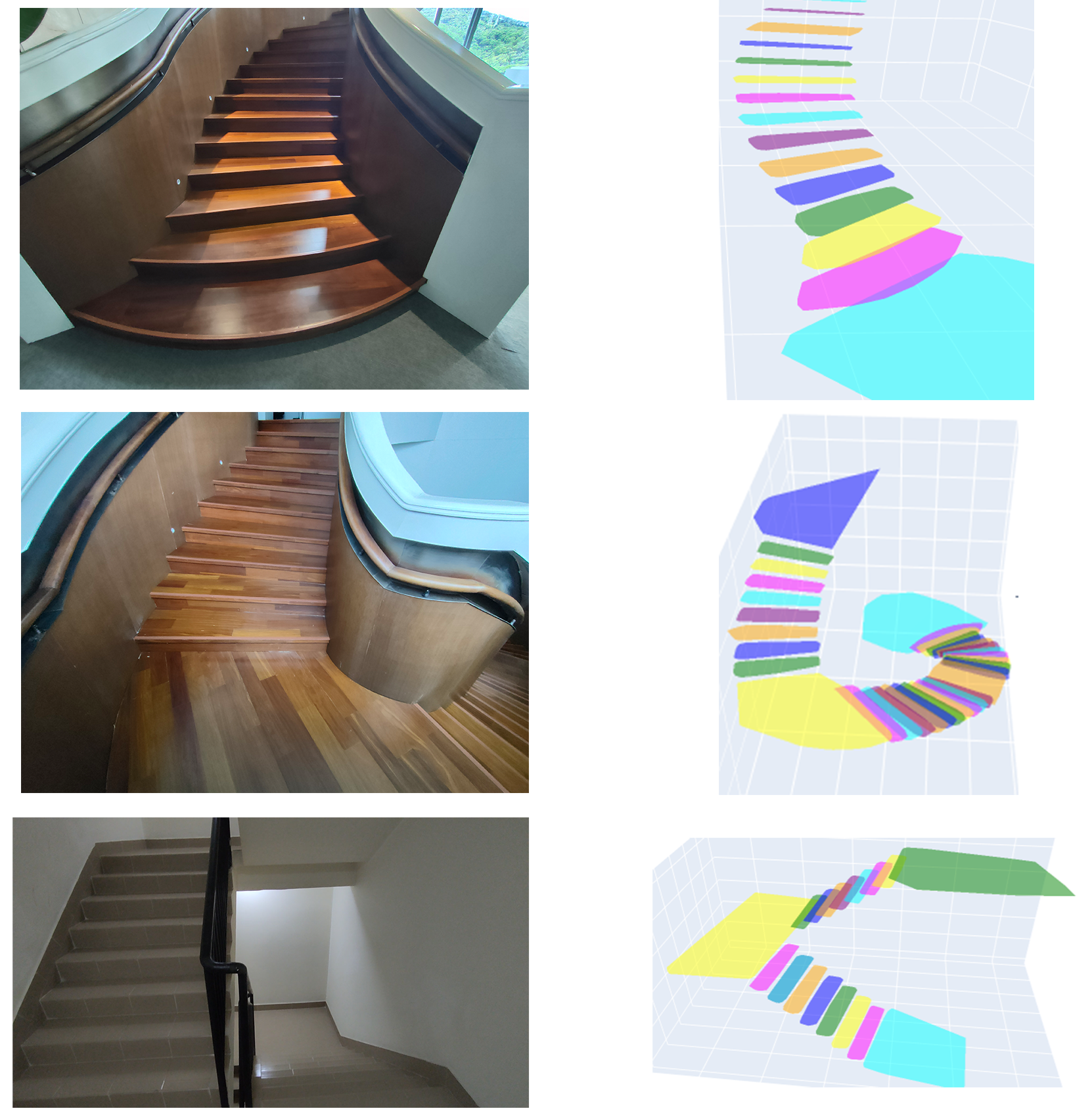}
\caption{Planar polygon semantic mapping results of spiral and straight stairs}
\label{fig:p0}
\end{figure}


\begin{figure*}[htbp]
\centering
\includegraphics[width=0.95\textwidth]{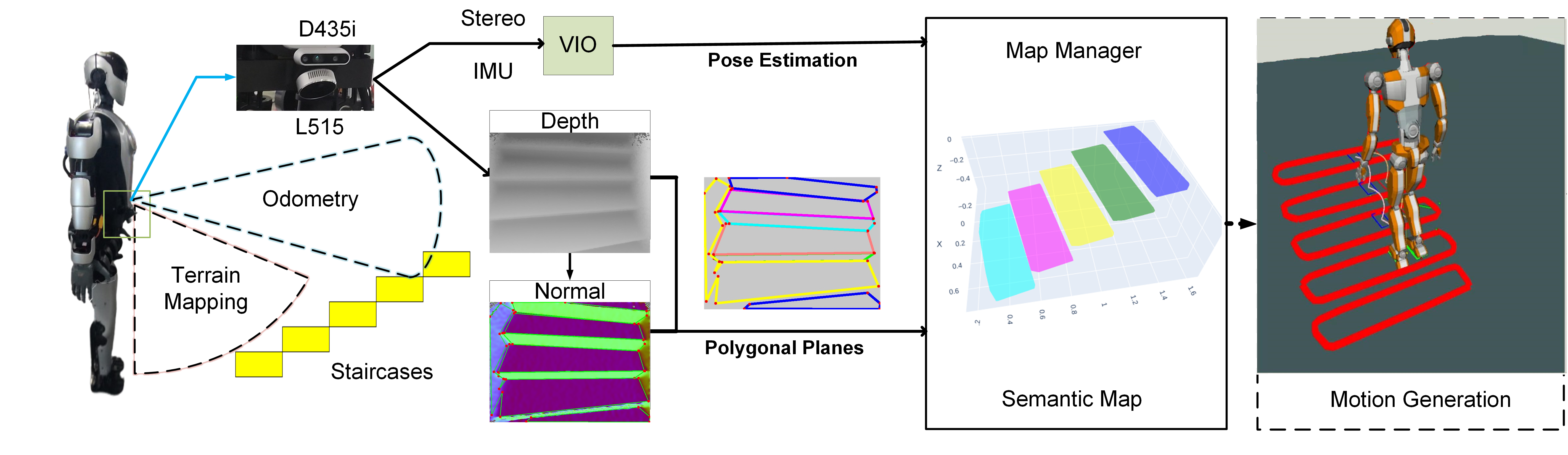}
\caption{Overview of the Planar Polygonal Semantic Mapping System Framework. The system inputs are depth images and robot pose estimates, which are processed to generate a terrain's polygonal planar semantic map. This map can be directly utilized to plan humanoid robot gaits.}
\label{fig:overview_of_system}
\vspace{-0.6cm}
\end{figure*}


However, current methods often overlook the impact of filtering techniques on extracting planes from raw data, which can lead to significant discrepancies in performance between noise-free simulated data and real depth camera data. While results showing an IOU of over 95\% in simulations drop below 80\% with real sensor data may be acceptable for quadruped robots, the requirement for larger stable areas for humanoid robots makes resolving this issue indispensable. The accuracy of the extracted plane results, including the planar normal vectors and centroid heights, is crucial as it directly influences the safety of the robot during its operation. Moreover, this thesis also considers real-time performance as a critical metric, placing computationally intensive and iterative image processing algorithms on the GPU to ensure that the processing time for each frame remains below the sensor cycle time. Issues such as dynamic obstacles and odometric drift also need to be addressed in map construction.

Our previous work \cite{stair} proposed using supervoxel-based plane segmentation for humanoid robot stairs. Supervoxel is a point cloud clustering method that is time-consuming. Thus, in this paper, we propose a system that performs real-time extraction of planar polygons and incremental semantic map management. This system first applies anisotropic diffusion filtering \cite{ADF} to significantly suppress noise in depth images, thereby enhancing the quality of depth images. It then uses edge detection algorithms to extract contours of planes from transformed normal vector images and simplifies these contours into polygons. Subsequently, the RANSAC algorithm fits the optimal plane equation for each polygonal region from the depth images. This process yields the results of polygonal plane extraction from single frames. To construct a global polygonal semantic map, the system integrates robot pose estimates obtained from odometry/SLAM, allowing newly observed polygons to be added to the map. During the polygon merging process, a simple method is employed to estimate and compensate for odometric drift in the vertical direction, ensuring more precise mapping results. Fig. \ref{fig:p0} illustrates the mapping results for different staircases. Building on this, we have also integrated the humanoid robot's gait planning algorithm with our mapping system, as shown in Fig. \ref{fig:overview_of_system}. The code can be accessed at \url{https://github.com/BTFrontier/polygon_mapping}.

\section{Related Works}\label{char2}

Accurate and efficient extraction of planar from sensory data is essential for autonomous navigation in complex environments. This section explores various methodologies to improve plane extraction, localization techniques, and their interconnections.

Region growing techniques such as those described in \cite{5} and \cite{6} evaluate inclusion based on the distance from points to a seed point and its normal plane. They merge planes based on the similarity of normal vectors and spatial distances, with \cite{6} utilizing Principal Component Analysis (PCA) to enhance resistance to noise by selecting seed points with low mean squared errors. However, these methods may exclude noisy and edge regions, impacting the accuracy of the extracted planes. Additionally, \cite{81} employs filtering to improve depth image quality but relies on a rough assumption that the depth value differences at plane edges are about $4~cm$, which limits its applicability for extracting accurate planes from depth images.

Graph-based methods and hierarchical clustering techniques effectively segment planar regions from point clouds. \cite{4} partitions point clouds into non-overlapping groups and uses Agglomerative Hierarchical Clustering (AHC) to merge nodes until a mean squared error threshold of plane fitting is exceeded.  Long et al. \cite{long} utilize the planar feature to deal with dynamic SLAM problems. 


The extraction of spatial planes and cylindrical surfaces for mapping and localization is addressed in \cite{3} and \cite{82}. \cite{3} refines cylindrical parameters using a nonlinear weighted least squares method and uses these surfaces for advanced localization. \cite{82} introduces a Plane-Edge-SLAM system that integrates planar and edge data for robust mapping and localization, addressing sensor noise with probabilistic plane fitting. 


Point cloud-based plane extraction techniques include the integration of RGB imagery with point clouds to enhance environmental feature detection. \cite{7} combines RGB images for staircase detection with point clouds for measuring step dimensions. \cite{8} reduces data volume and noise through voxel grid filtering, which is beneficial for subsequent region growing and PCA applications. \cite{11} proposes a novel approach combining octree and nearest-neighbour searches to extract planar regions from point clouds, although challenges remain in processing the data in real-time.

Advanced terrain mapping and environmental modelling are crucial for navigation and path planning. \cite{17} introduces a novel terrain mapping method that incorporates state estimation drift and sensor noise into its probabilistic terrain estimates based solely on kinematics and inertial measurements. Building on this, \cite{18} converts point clouds into grid-based elevation maps using GPU acceleration for dynamic obstacle elimination.  \cite{zhang} proposed to use planar semantics for visual SLAM relocation after a humanoid robot falls.




\begin{figure}[htbp]
\centering
\includegraphics[width=0.85\columnwidth]{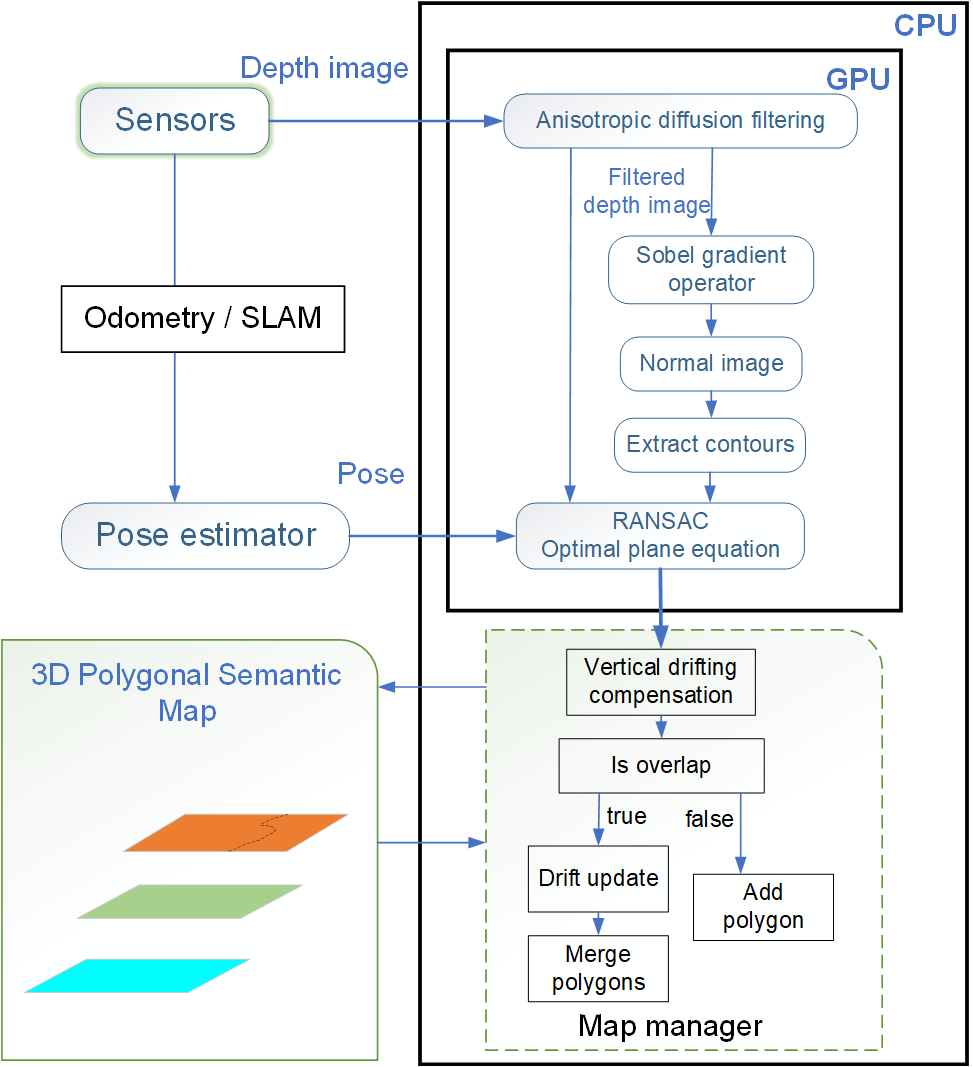}
\caption{Overview of processing.
} 
\label{fig:overview_of_processing}
\vspace{-0.5cm}
\end{figure}



\section{\textsc{System Framework And Methods}}\label{char3}


This section delineates the architecture and methodologies employed in our system for real-time planar semantic mapping tailored for humanoid robots, aimed at achieving high accuracy and efficiency in dynamic and complex environments.

\subsection{Overview}
As illustrated in Fig. \ref{fig:overview_of_processing}, the system inputs are depth images from depth sensors and their pose estimates, which can be obtained through odometry or SLAM. Initially, the depth images are transferred to GPU memory, where they undergo anisotropic diffusion filtering. Subsequently, the Sobel operator computes the normal vector images, and the Canny detector identifies edges to extract contours. Simplifying these contours yields polygonal segmentation results for each plane. Combined with the segmented polygons and depth images, the RANSAC algorithm fits the optimal plane model for each polygon. Utilizing the pose estimates, newly obtained polygons are merged with the existing semantic polygonal map, incorporating a vertical drift compensation during this process.

\begin{figure}[htbp]
\centering
\includegraphics[width=\columnwidth]{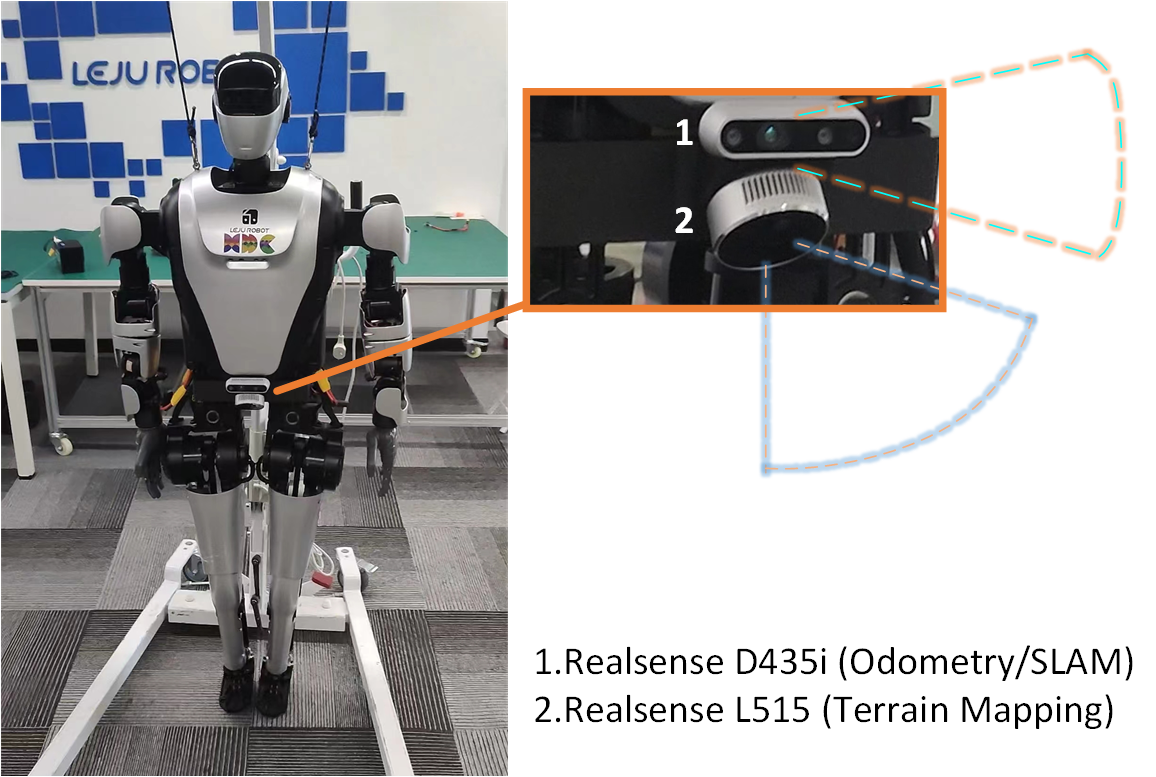}
\caption{Multi-sensory setup installed at the waist of the Leju humanoid robot \cite{leju-robot}. } 
\label{fig:sensory_devices}
\vspace{-0.2cm}
\end{figure}
\subsection{System Sensory Equipment}
A sensory setup as depicted in Fig. \ref{fig:sensory_devices} equips the robot for navigation and mapping in complex environments. The Realsense L515 sensor, oriented towards the ground, captures depth data even in low-light conditions due to its LiDAR technology, though it is less effective on light-absorbing surfaces. For odometric pose estimation, a Realsense D435i utilizes its stereo infrared cameras and IMU. It is critical to disable the infrared speckle projector during operation to prevent interference with visual odometry. The system employs VINS Fusion \cite{vinsfusion} for odometry, optimizing for both computational efficiency and stability.

\subsection{Anisotropic Diffusion and Normal Vector Calculation}
This section discusses the implementation of anisotropic diffusion for noise reduction in depth images and the subsequent calculation of normal vectors using the Sobel operator and camera intrinsic parameters.

\begin{figure}[htbp]
	\centering

        \begin{minipage}[b]{0.13\textwidth}
		\centering
		\includegraphics[width=\textwidth]{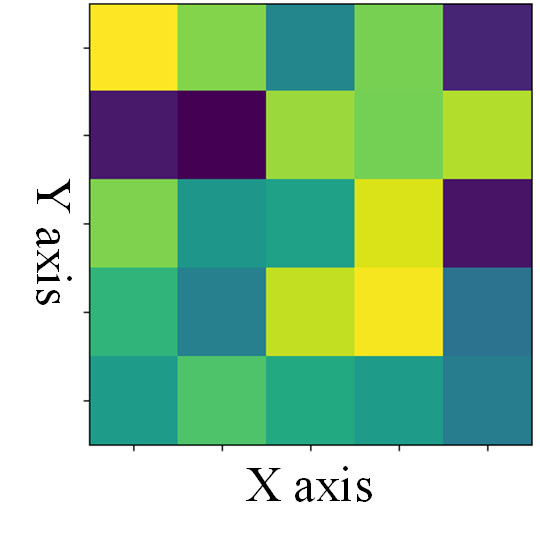}
		\subcaption{raw image}
		\label{fig_3_1_0d}
	\end{minipage} 
     \begin{minipage}[b]{0.18\textwidth}
		\centering
		\includegraphics[width=\textwidth]{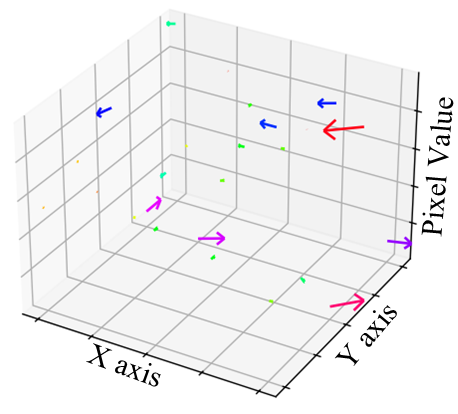}
		\subcaption{gradient}
		\label{fig_3_1_0b}
	\end{minipage} 
        \begin{minipage}[b]{0.125\textwidth}
		\centering
		\includegraphics[width=\textwidth]{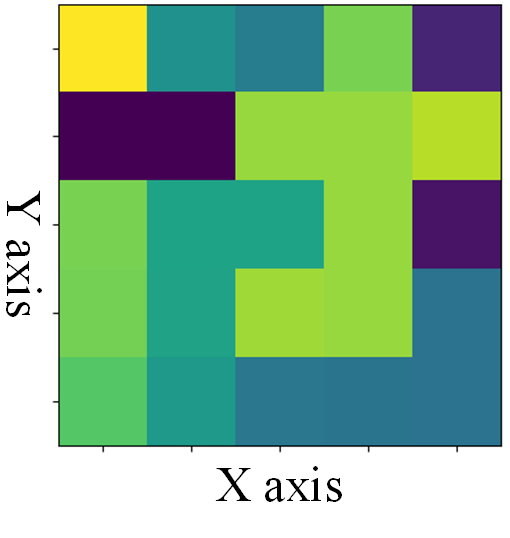}
		\subcaption{filtered image}
		\label{fig_3_1_0f}
	\end{minipage} 
	\caption{Anisotropic Diffusion Filtering}
	\label{fig:ADF_theoretical effects}
 \vspace{-0.3cm}
\end{figure}

\textbf{Anisotropic Diffusion Filtering:}
Anisotropic diffusion filtering is employed to enhance image quality by reducing noise while preserving significant edges \cite{ADF}. This filtering is particularly effective for depth images where noise can significantly affect the accuracy of gradient-based calculations.

\begin{equation}
\frac{\partial I}{\partial t} = \nabla \cdot (c(\|\nabla I\|) \nabla I)
\end{equation}
where \( I \) is the image intensity, \( \nabla I \) denotes the image gradient, and \( c(\|\nabla I\|) \) is the diffusion coefficient, which decreases with increasing gradient magnitude to preserve edges while smoothing other areas.

\begin{figure}[htbp]
	\centering
	\begin{minipage}[c]{0.23\textwidth}
		\centering
		\includegraphics[width=\textwidth]{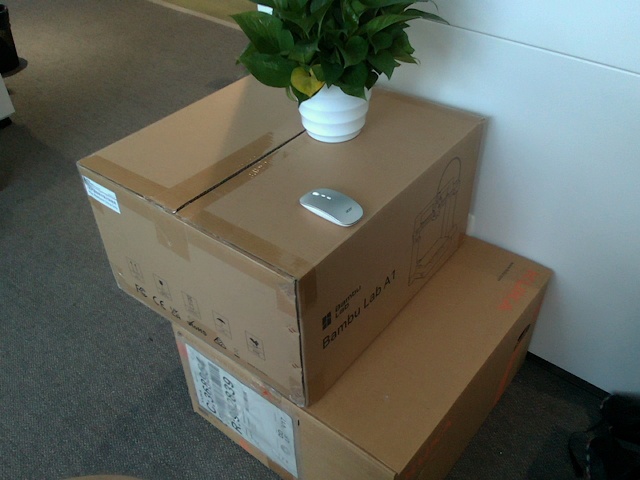}
		\subcaption{ }
		\label{fig_2_1_2a}
	\end{minipage} 
	\begin{minipage}[c]{0.23\textwidth}
		\centering
		\includegraphics[width=\textwidth]{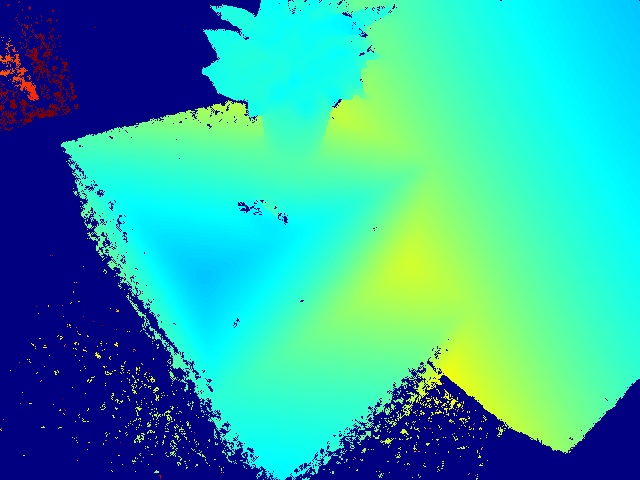}
		\subcaption{ }
		\label{fig_2_1_2b}
	\end{minipage} \\
        \begin{minipage}[c]{0.23\textwidth}
		\centering
		\includegraphics[width=\textwidth]{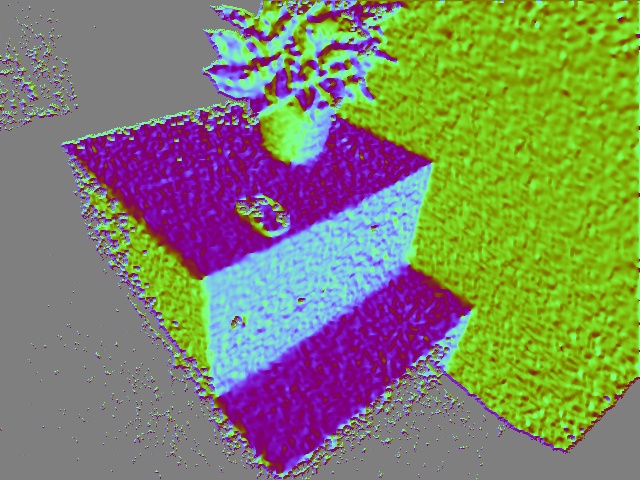}
		\subcaption{ }
		\label{fig_2_1_2c}
	\end{minipage} 
	\begin{minipage}[c]{0.23\textwidth}
		\centering
		\includegraphics[width=\textwidth]{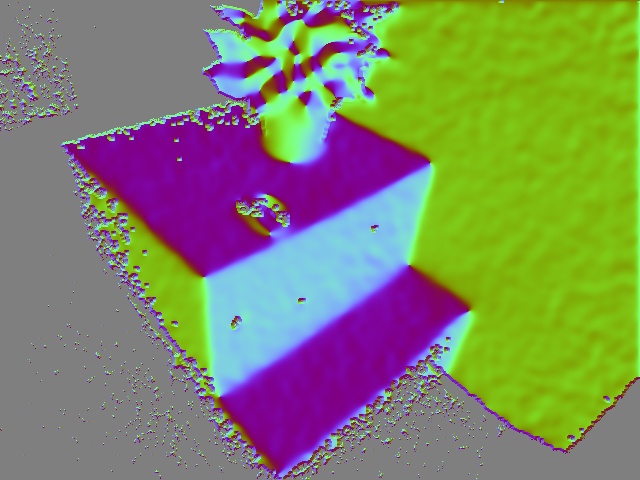}
		\subcaption{ }
		\label{fig_2_1_2d}
	\end{minipage} 
	\caption{(a) RGB image, (b) Depth image, (c) Normal vector image calculated from the depth image before filtering, and (d) Normal vector image calculated from the depth image after filtering.}
	\label{fig:ADF_ex}
 \vspace{-0.2cm}
\end{figure}

Fig. \ref{fig:ADF_theoretical effects} illustrates the theoretical effects of anisotropic diffusion filtering. 

\textbf{Normal Vector Calculation:}
Following the diffusion process, normal vectors at each pixel are calculated using the Sobel operator to detect gradients and the camera's intrinsic matrix to project these gradients into 3D space.

\begin{equation}
\mathbf{n}_p = -K^{-1} \begin{bmatrix} G_x \\ G_y \\ 1 \end{bmatrix}
\end{equation}
where \( G_x \) and \( G_y \) are the gradients computed using the Sobel operator, \( K \) is the camera intrinsic matrix, and \( \mathbf{n}_p \) is the normal vector at pixel \( p \).

As shown in Fig. \ref{fig:ADF_ex}, due to sensor noise, the gradient in the original depth image exhibits significant jumps within planar areas. After filtering the depth image, the normal vector image becomes much smoother, while still retaining critical edge details. This enhancement facilitates the extraction of more complete and continuous planar regions, essential for accurate mapping and robot navigation.

\begin{algorithm}
\caption{GPU-based Anisotropic Diffusion and Normal Vector Calculation}
\begin{algorithmic}[1]
\Require Depth image $I$ , parameters $\gamma$, $k$, iterations $N$, Camera intrinsic matrix $K$
\Ensure Smoothed image $I_{smooth}$, Normal vectors $N_p$
\State Initialize $I_{smooth} \gets I$
\For{$i = 1$ to $N$}
    \For{each pixel $p$ in $I_{smooth}$ \textbf{parallel}}
        \State Compute $\nabla I$ at $p$
        \State $c_p \gets \exp\left(-\left(\frac{\|\nabla I_p\|}{k}\right)^2\right)$
        \State Update $I_{smooth, p} \gets I_{smooth, p} + \gamma \cdot c_p \cdot \nabla^2 I_p$
    \EndFor
\EndFor
\For{each pixel $p$ in $I_{smooth}$ \textbf{parallel}}
    \State Compute gradients $G_x, G_y$ using Sobel operator at $p$
    \State Calculate normal vector $N_p \gets -K^{-1} \begin{bmatrix} G_x \\ G_y \\ 1 \end{bmatrix}$
    \State Normalize $N_p$
\EndFor
\end{algorithmic}
\end{algorithm}

The pseudocode for GPU-accelerated anisotropic diffusion filtering and normal vector calculation of depth images is presented in \textbf{Algorithm 1}.

\subsection{Plane Segmentation}
This section details the process of plane segmentation using the normal vectors calculated from the depth images. The algorithm involves edge detection using the Canny method, contour extraction, and simplification of these contours into polygonal shapes. Planar models are then fitted to these polygons using the RANSAC algorithm.

\begin{figure}[htbp]
	\centering
	\begin{minipage}[c]{0.23\textwidth}
		\centering
		\includegraphics[width=\textwidth]{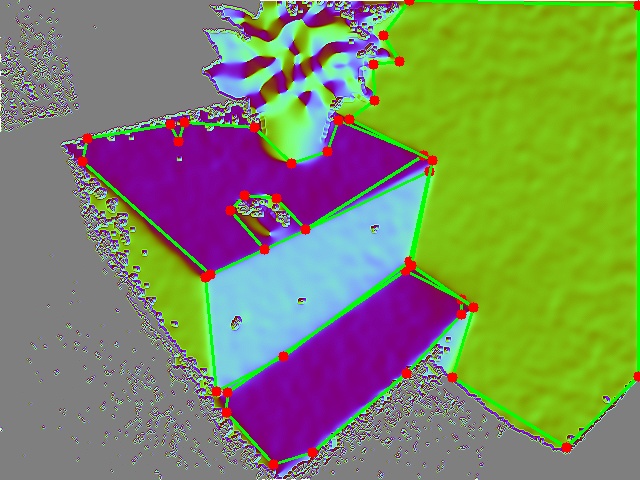}
		\subcaption{ }
		\label{fig_3_2_1}
	\end{minipage} 
	\begin{minipage}[c]{0.23\textwidth}
		\centering
		\includegraphics[width=\textwidth]{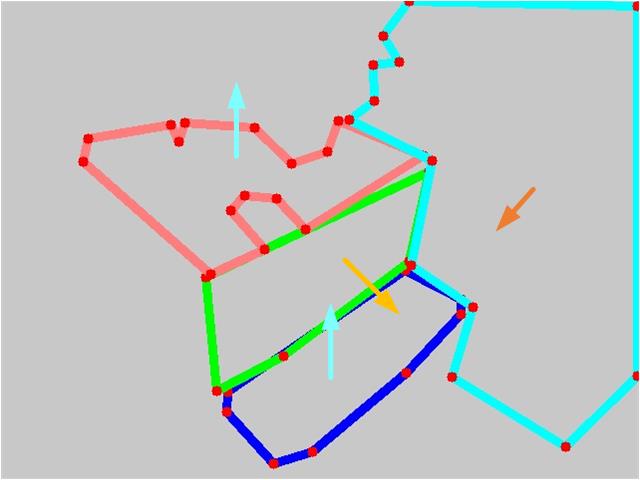}
		\subcaption{ }
		\label{fig_3_2_2}
	\end{minipage} 

	\caption{Illustration of the plane extraction and fitting process: (a) shows the polygonal contours of planes extracted from the normal vector image; (b) depicts the optimal plane model parameters fitted using the RANSAC algorithm.}
	\label{fig:Extracting non-convex polygons}
\vspace{-0.3cm}
\end{figure}


\textbf{Edge Detection and Contour Extraction:}
Initially, edges are detected from the normal vector image using the Canny edge detection algorithm. Subsequently, contours are extracted from these edges, and these contours are simplified into polygons that represent potential planar regions.

\textbf{RANSAC for Plane Fitting:}
The polygons are then projected into 3D space using the camera's intrinsic parameters, and these points are used to fit planes using the RANSAC algorithm. This process iterates until the maximum iterations are reached or a satisfactory model is found.

As shown in Fig. \ref{fig:Extracting non-convex polygons}, the image demonstrates the results of extracting polygonal planes and fitting plane models in scenarios where obstacles obscure the view. Notably, the largest non-convex polygon is identified as the main plane of the cardboard box, despite being partially obscured by the obstacle.

The computations, particularly the intensive RANSAC algorithm for plane fitting, are accelerated using CUDA:
\begin{equation}
\text{Minimize } \sum_{i=1}^{N} \rho(\mathbf{p}_i, \pi)
\end{equation}
where \(\rho\) is a distance function measuring the inliers to the plane \(\pi\) and \(\mathbf{p}_i\) are the points considered by the RANSAC algorithm. This GPU acceleration ensures that our system achieves real-time performance even when processing extensive data from complex scenes.

\vspace{-0.2cm}
\begin{algorithm}
\setstretch{0.9} 
\captionsetup{skip=1pt} 
\caption{GPU-accelerated Plane Extraction and Fitting}
\begin{algorithmic}[1]
\Require Depth image $I$, Polygon contours $C$, Camera intrinsic matrix $K$, RANSAC parameters
\Ensure Plane models and their parameters

\For{each detected contour $c$}
    \State Extract depth values within $c$ from $I$ 
    \State Convert depth values to 3D points $P$ using  $K$
    \State Initialize best model, best error , best inliers = infinity
    \For{iterations}
        \State Select random subset of 3 points
        \State Fit plane model to these points
        \State error = 0, inliers = 0
        \For{each point in $P$ \textbf{parallel}}
            \State $d, is\_inlier \gets $d$, ($d \textless threshold$)$
            \State error += $d$
            \State inliers += ($is\_inlier$ ? 1 : 0)
        \EndFor
        \If{ (error \textless best error)}
            \State best model, best error  $\gets$ current model, error 
            \State  best inliers $\gets$ inliers
        \EndIf
    \EndFor
    \If{($best inliers/total points \textgreater 0.9$) }
        \State Store the model and parameters
    \EndIf
\EndFor
\end{algorithmic}
\end{algorithm}
\vspace{-0.3cm}

The pseudocode for GPU-accelerated plane extraction and fitting is presented in \textbf{Algorithm 2}. This algorithm efficiently identifies and models planar regions within the image, discarding any models where the fraction of outliers exceeds 10\%. The use of GPU acceleration ensures that the computations are performed swiftly, facilitating real-time processing capabilities.

\subsection{Polygons Integration and Drift Compensation}
Due to the limited field of view of sensors, a single plane may need to be observed multiple times from different positions to capture its complete shape. This section describes the management of these observations within a map manager, which integrates multiple polygonal observations, compensates for vertical drift, and merges polygons when appropriate.



\textbf{Polygons Integration:}
Merging is considered if the vertical distance between the centroids of the polygons is within a specified drift tolerance. When two polygons are identified as overlapping and the vertical height difference between their centroids is within the allowable range, the current vertical drift is updated and the new polygon is adjusted accordingly before merging. The merging conditions are as follows:
\begin{equation}
\Delta z = |z_{centroid, new} - z_{centroid}|
\end{equation}
\begin{equation}
\text{Merge if } \Delta z \leq \textit{Drift~Tolerance}
\end{equation}
where the drift tolerance is assumed to be $5~cm$. This approach simplifies the integration and management of new data into the map, enhancing the system's ability to maintain accurate and reliable representations of the environment.

\begin{figure}[htbp]
\centering
\includegraphics[width=0.83\columnwidth]{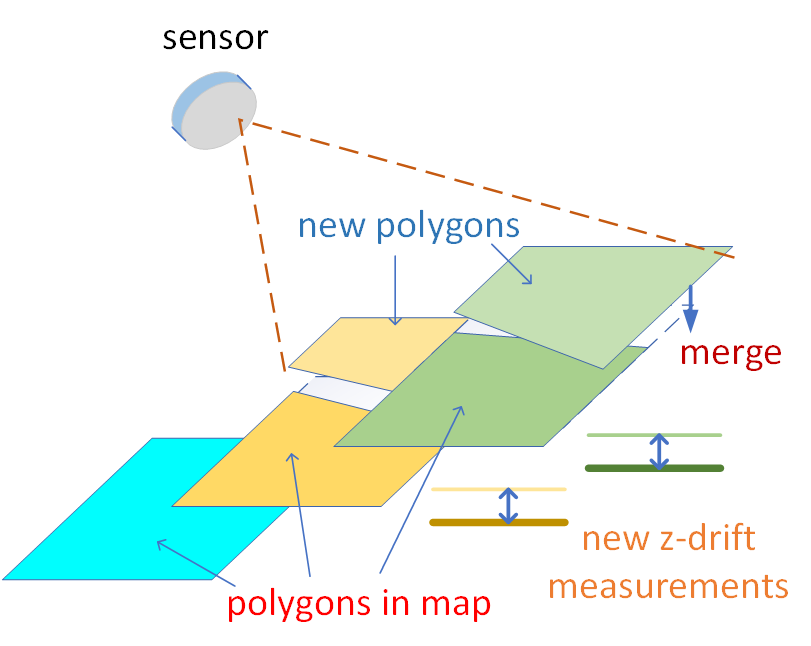}
\caption{Schematic representation of map merging and drift compensation. This diagram illustrates the process of integrating newly observed polygons with the existing map while accounting for vertical drift to ensure accurate alignment and positional integrity.}
\label{fig:map_merge_compensation}
\vspace{-0.3cm}
\end{figure}

\textbf{Vertical Drift Compensation:}
The Kalman filter process for vertical drift compensation includes the following steps:
\begin{equation}
\hat{x}_{k|k-1} = \hat{x}_{k-1|k-1}   
\end{equation}
\begin{equation}
P_{k|k-1} = P_{k-1|k-1} + \sigma_{p}  
\end{equation}
\begin{equation}
K_k = \frac{P_{k|k-1}}{P_{k|k-1} + \sigma_{m}}  
\end{equation}
\begin{equation}
\hat{x}_{k|k} = \hat{x}_{k|k-1} + K_k(z_k - \hat{x}_{k|k-1})  
\end{equation}
\begin{equation}
P_{k|k} = (1 - K_k)P_{k|k-1}  
\end{equation}

\textit{Where:}
\begin{itemize}
    \item $\hat{x}_{k|k-1}$: Predicted state estimate of the vertical drift before the new measurement.
    \item $P_{k|k-1}$: Predicted covariance, indicating the uncertainty before updating.
    \item $K_k$: Kalman Gain, determining the weight given to the new measurement versus the prediction.
    \item $z_k$: New measurement of drift, calculated as the average of the vertical component of centroid distances for all detected polygons versus the map.
    \item $\hat{x}_{k|k}$: Updated drift estimate incorporating the new measurement.
    \item $P_{k|k}$: Updated covariance, indicating the reduced uncertainty post-measurement.
\end{itemize}

In cases with substantial uncertainty in vertical drift, such as significant IMU noise, it is advisable to greatly increase $\sigma_p$ relative to $\sigma_m$ to ensure measurements predominantly influence the updates.
This method ensures that each new measurement refines the drift estimate, allowing for precise adjustments over time, which is crucial for maintaining map accuracy and robot navigation reliability.

As shown in Fig. \ref{fig:map_merge_compensation}, each new polygon obtained per frame is transformed into the map coordinate system for overlap assessment with existing polygons. Following vertical drift compensation, these polygons are either added to the map or merged with existing polygons to maintain accurate and consistent mapping.




\section{\textsc{Experiments Results and Evaluations}}
\label{char4}


This section of the paper presents a comprehensive evaluation of the developed algorithm for real-time planar semantic mapping of stairs for humanoid robots. The experiments are structured into three distinct parts to validate the accuracy, real-time performance, and practical utility of the algorithm under varied conditions and resolutions. The experiments were conducted using a laptop with an Intel i9 13900HX processor and an NVIDIA RTX 4060 graphics card, running Ubuntu 20.04.

\subsection{Single-Frame Mapping Accuracy }
The first part of the experiments focuses on assessing the single-frame mapping accuracy using a standard-sized staircase model. This involves:

Presentation of mapping results including photographs of the stairs, normal vector images, and 3D representations in the world coordinate system.

\begin{figure}[htbp]
	\centering
	\begin{minipage}[c]{0.23\textwidth}
		\centering
		\includegraphics[width=\textwidth]{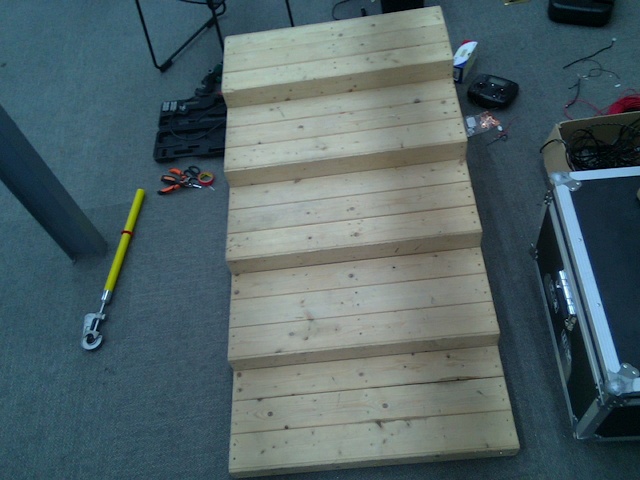}
		\subcaption{Stair model}
		\label{fig:Stair model}
	\end{minipage} 
	\begin{minipage}[c]{0.23\textwidth}
		\centering
		\includegraphics[width=\textwidth]{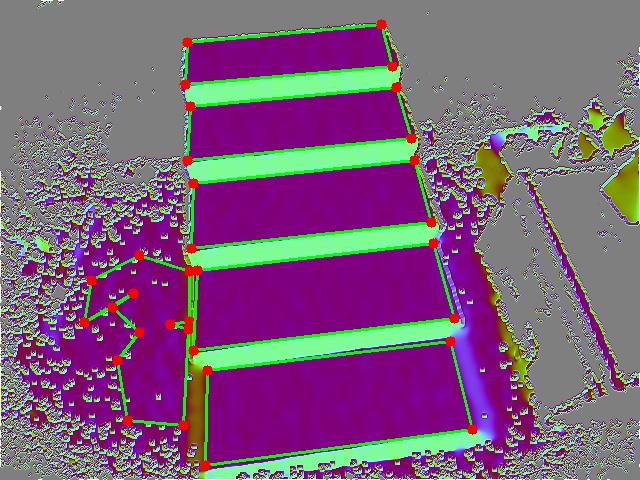}
		\subcaption{Extract polygons}
		\label{fig:Stair model Extract polygons}
	\end{minipage} \\

	\caption{Extract the polygonal planes effects of the entire stair model in a single frame image.}
	\label{fig:ex1a}
\vspace{-0.3cm}
\end{figure}

\begin{figure}[htbp]
\centering
\includegraphics[width=0.6 \columnwidth]{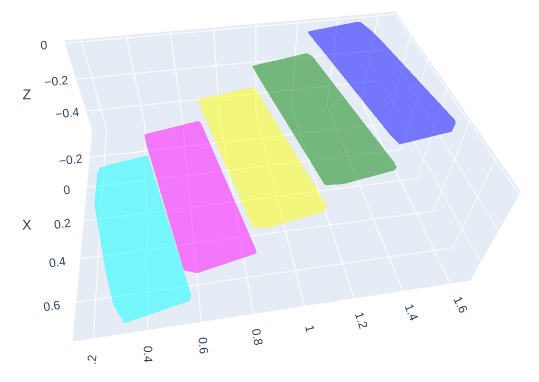}
\caption{3D polygonal semantic map of the stair model}
\label{fig:ex1b}
\vspace{-0.3cm}
\end{figure}

Fig. \ref{fig:ex1a} displays the staircase model photo alongside the process of single-frame plane extraction. Subsequently, Fig. \ref{fig:ex1b} shows the mapping results following the single-frame plane extraction. Quantitative analysis involves the comparison of the effective area covered in the maps to the actual area, the mean and standard deviation of the stair normal vector angles, and the height error for each stair set.

\begin{table}[ht]
\centering
\caption{Results of Single-Frame Mapping Accuracy Tests}
\label{tab:mapping_accuracy}
\begin{tabular}{cccccc}
\hline
\textbf{Test \#} & \textbf{$\alpha$ ($^\circ$)} & \textbf{$\sigma_\alpha$($^\circ$)} & \textbf{IOU} & \textbf{$\Delta d(mm)$} & \textbf{$\sigma_{\Delta d}(mm)$} \\
\hline
1 & 2.22 & 0.457 & 0.936 & 3.2 & 1.51 \\
2 & 2.15 & 0.410 & 0.922 & 1.8 & 1.72 \\
3 & 2.22 & 0.453 & 0.940 & 1.3 & 2.03 \\
\hline
\end{tabular}
\smallskip \\
\footnotesize
 \vspace{-0.3cm}
\end{table}

In Table I, the analysis of the single-frame plane extraction results for the staircase model reveals that the algorithm achieves an average angle $\alpha$ between the step normal vectors and the direction of gravitational acceleration of approximately $2.2^\circ$, with a standard deviation $\sigma_\alpha$ of $0.44^\circ$. The average error in step height $\Delta d$ is about $2.1$ mm, with a standard deviation also of $2.1$ mm, and the Intersection over Union (IOU) for the extracted areas is around $0.93$. These results indicate that the algorithm exhibits very high single-frame repeatability, with plane extraction errors generally on the same order of magnitude as sensor measurement errors.


\begin{table}[ht]
\setlength{\tabcolsep}{1mm}
\centering
\caption{Comparison of Plane Extraction Performance}
\label{tab:mapping_comparison}
\begin{tabular}{cccccc}
\hline
\textbf{Method} & \textbf{$\Delta d(mm)$} & \textbf{$\sigma_{\Delta d}(mm)$} & \textbf{$\sigma_\alpha$($^\circ$)} & \textbf{IOU} & \textbf{Runtime} \\
\hline
PPRCoRT & 20.2 & 5.33 & 0.48 & 0.88 & 24 ms\\
\textbf{Ours} & 2.1 & 1.75 & 0.44 & 0.93 & 7.5 ms\\
\hline
\end{tabular}
\smallskip \\
\footnotesize
 \vspace{-0.3cm}
\end{table}

The average results of the single-frame mapping for the staircase model are compared with those obtained by the PPRCoRT method \cite{6}, as presented in Table II. The PPRCoRT results are derived from merging outcomes of three frames, hence the processing time reference is based on their reported duration of $8$ ms per frame for a $640 \times 480$ resolution depth image. It is evident that, except for the standard deviation of the normal vector, our method outperforms PPRCoRT across all other metrics. This indicates that our single-frame plane extraction not only maintains high accuracy but also demonstrates superior performance in critical aspects of semantic mapping.

\subsection{Performance Evaluation of Algorithmic Components}
The second experiment evaluates the runtime efficiency of key components of the algorithm, namely the anisotropic diffusion and RANSAC plane extraction processes.
The parameters used for testing real-time performance included 10 iterations of anisotropic diffusion filtering and 10 iterations of RANSAC. These settings were sufficient to reduce noise in the depth images significantly and to extract planes quickly and accurately. However, the number of filtering iterations can be increased to over 20 in higher noise levels.

\begin{figure}[htbp]
	\centering
	\begin{minipage}[c]{0.4\textwidth}
		\centering
		\includegraphics[width=\textwidth]{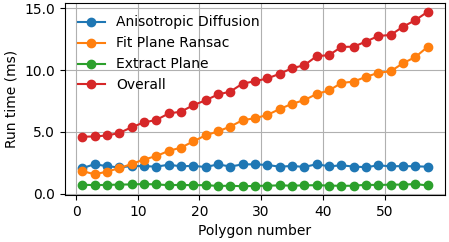}
		\subcaption{320×240 Performance curves}
		\label{fig:ex2 320 240}
	\end{minipage} \\
	\begin{minipage}[c]{0.4\textwidth}
		\centering
		\includegraphics[width=\textwidth]{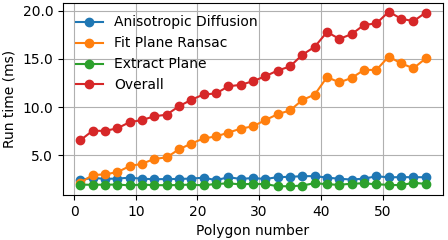}
		\subcaption{640×480 Performance curves}
		\label{fig:ex2 640 480}
	\end{minipage} \\
	\caption{Single frame running time curve of the program at different resolutions}
	\label{fig:ex2a}
  \vspace{-0.3cm}
\end{figure}

As demonstrated in Fig. \ref{fig:ex2a}, the processing time for parallel operations of anisotropic diffusion and RANSAC remains relatively stable even as image resolution increases. However, there is a linear increase in RANSAC processing time with the number of polygons processed per frame. This increase occurs because the code sequentially performs RANSAC fitting for each polygon. Nevertheless, in practical scenarios, the number of valid polygons extracted from a single-frame image is generally small, making this increase negligible. It is evident from the results that under normal operating conditions, the algorithm consistently maintains processing times below $15~ms$ for both resolutions examined.

\subsection{Comparative Analysis with Elevation Mapping}
In the third part of the experiments, we compare our method with the elevation mapping cupy \cite{18} (hereafter referred to as EM\_cupy), focusing on comprehensive mapping experiments conducted on a straight staircase. The comparison primarily evaluates the accuracy of plane extraction, Intersection over Union (IOU), step height errors, and the overall representation of the map vertical drift compensation. Further, we analyze several distinct advantages of our method when implemented on humanoid robots. 
In the comparative experiments, the EM\_cupy method was configured with a resolution of $1~cm$. Additionally, default settings in the algorithm, such as outlier removal and filtering, were maintained unchanged to ensure consistency in performance evaluation. Both methods employed VINS Fusion \cite{vinsfusion} as the odometry input, ensuring a standardized basis for comparison in terms of data integration and positional accuracy.

\begin{figure}[htbp]
	\centering
	\begin{minipage}[b]{0.32\textwidth}
		\centering
		\includegraphics[width=\textwidth]{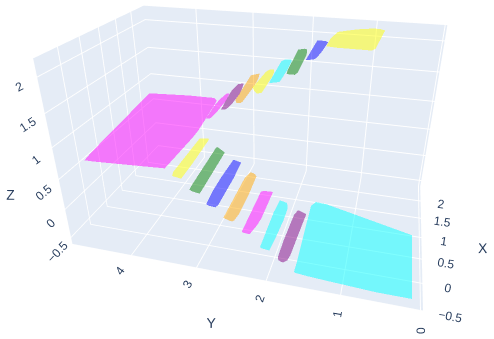}
		\subcaption{ }
		\label{fig_2_1_2a}
	\end{minipage} 
	\begin{minipage}[b]{0.12\textwidth}
		\centering
		\includegraphics[width=\textwidth]{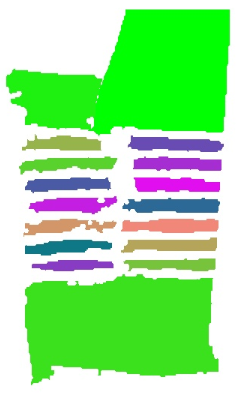}
		\subcaption{ }
		\label{fig_2_1_2b}
	\end{minipage} \\
        \begin{minipage}[b]{0.32\textwidth}
		\centering
		\includegraphics[width=\textwidth]{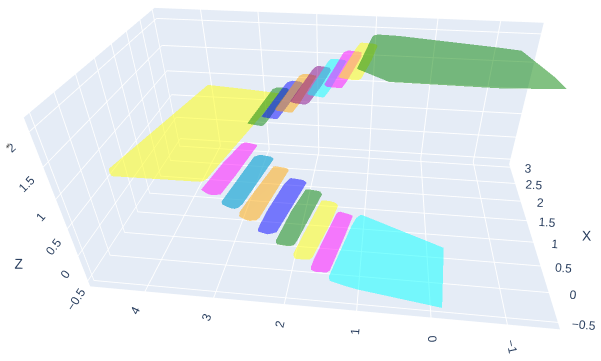}
		\subcaption{ }
		\label{fig_2_1_2c}
	\end{minipage} 
	\begin{minipage}[b]{0.12\textwidth}
		\centering
		\includegraphics[width=\textwidth]{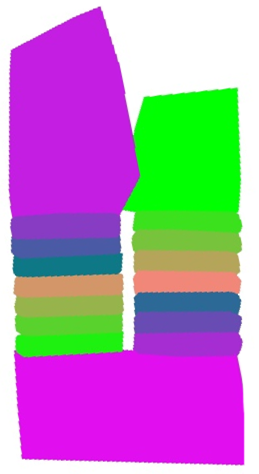}
		\subcaption{ }
		\label{fig_2_1_2d}
	\end{minipage} 
	\caption{Comparison of 3D and 2D mapping results by EM\_cupy and our method. Subfigures (a) and (b) depict the 3D and 2D maps from EM\_cupy, respectively. Subfigures (c) and (d) show the 3D and 2D maps from our method, respectively.}
	\label{fig:ex3a}
 \vspace{-0.3cm}
\end{figure}

As illustrated in Fig. \ref{fig:ex3a}, it is evident that our method achieves a greater effective planar area extraction compared to EM\_cupy. As shown in Table III, a comparison of plane extraction results for the staircase between the two methods reveals that both approaches achieve height errors of centroids of the stair levels within millimetres after vertical drift compensation. Such precision is crucial for the safety of humanoid robot motion. Additionally, our method surpasses the EM\_cupy approach regarding normal vector accuracy and Intersection over Union (IOU) metrics. Moreover, our process is not limited by resolution constraints. Given that EM\_cupy takes over ten seconds to extract planes at a $1~cm$ resolution in our experiments, real-time performance is not compared here.

\begin{table}[ht]
\centering
\caption{Comparative Analysis of Mapping Methods}
\label{tab:mapping_comparison}
\begin{tabular}{cccccc}
\hline
\textbf{Method} & \textbf{Map Type} & \textbf{$\alpha$ ($^\circ$)} & \textbf{$\sigma_\alpha$ ($^\circ$)} & \textbf{IOU} & \textbf{$\Delta d$} \\
\hline
y & Grid (1 cm) & 8.17 & 4.39 & 0.61 & 5.7 \\
\textbf{Ours} & Continuous & 4.74 & 1.02 & 0.96 & 5.3  \\
\hline
\end{tabular}
\smallskip \\
\footnotesize
\end{table}

\begin{figure}[htbp]
	\centering
	\begin{minipage}[b]{0.23\textwidth}
		\centering
		\includegraphics[width=\textwidth,height=3cm]{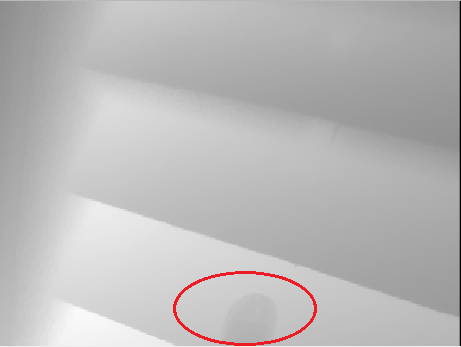} 
		\subcaption{Robot foot}
		\label{fig_2_1_2a}
	\end{minipage} 
	\begin{minipage}[b]{0.23\textwidth}
		\centering
		\includegraphics[width=\textwidth,height=3cm]{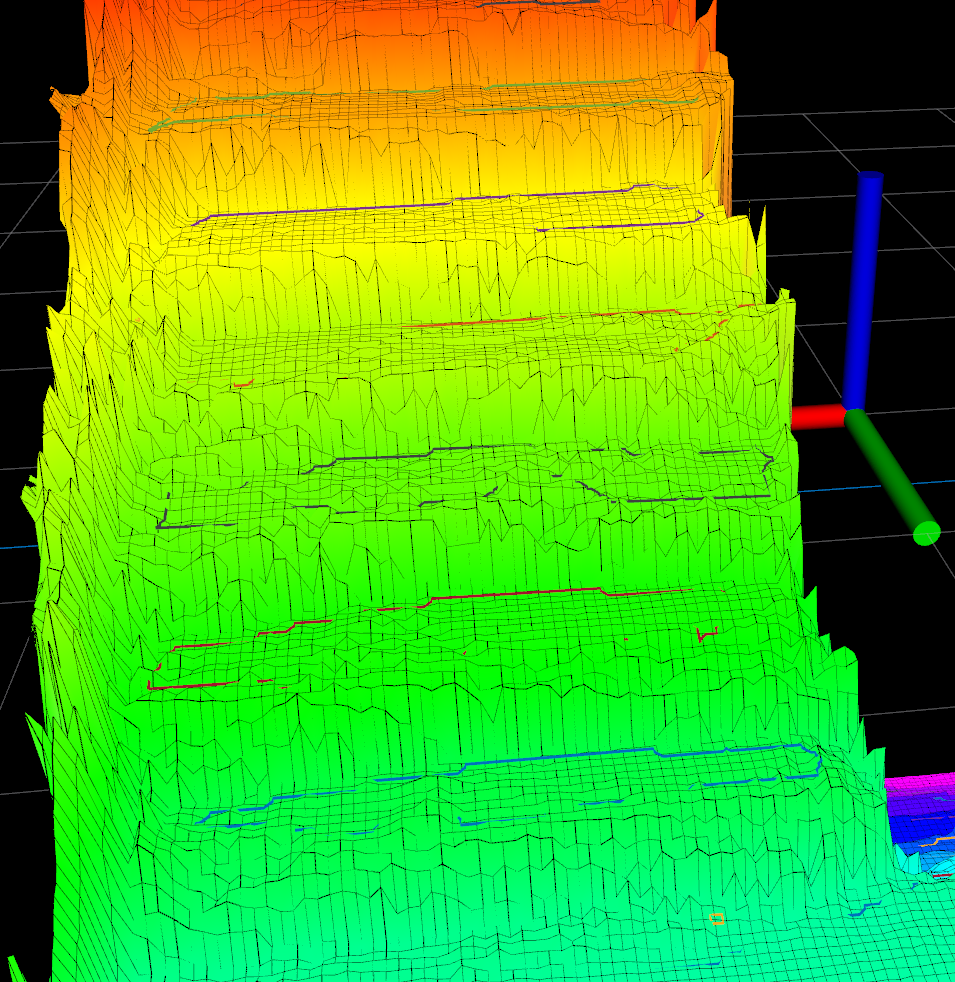} 
		\subcaption{Maps built are affected}
		\label{fig_2_1_2b}
	\end{minipage} \\
   \vspace{-0.1cm}
	\caption{The point cloud-based method does not immediately identify and remove dynamic obstacles, resulting in the retention of interference information in the map.}
	\label{fig:ex3z}
\end{figure}

During the mapping process using both methods, it was observed that although EM\_cupy incorporates an Overlap clearance mechanism, it struggles in specific scenarios, such as when a robot climbs stairs. As shown in Fig. \ref{fig:ex3z}, EM\_cupy may inaccurately place the robot's front foot in the map, whereas our method preemptively removes non-planar areas during the mapping process, thereby exhibiting better interference resistance. Furthermore, our method's effective area of plane extraction significantly exceeds that of EM\_cupy, presenting distinct advantages for motion planning in humanoid robots. In the next section, we will demonstrate the effectiveness of our mapping approach through a simulation of gait planning for humanoid robots using the maps generated by our method.

\subsection{Motion Planning Simulation}
In this section, we utilize the semantic map of polygonal planes developed from a straight staircase to conduct motion planning simulations with a humanoid robot, thereby validating the practicality of our plane extraction method. 
\begin{figure}[htbp]
\centering
\includegraphics[width= \columnwidth ]{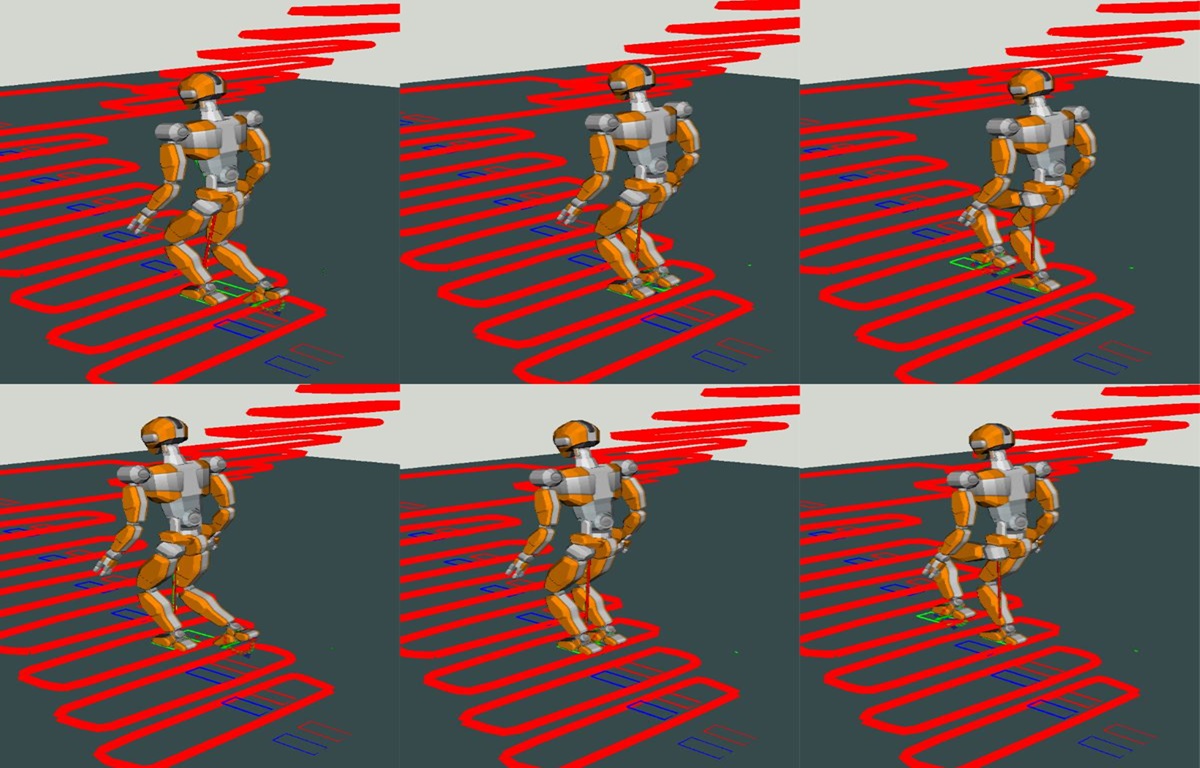}
\caption{Simulation of humanoid robot gait planning. }
\label{fig:ex4}
 \vspace{-0.3cm}
\end{figure}

As shown in Fig. \ref{fig:ex4}, the humanoid robot starts its stair-climbing simulation with both feet aligned before the staircase. The motion is broken down into sequential steps. After planning the footsteps, a walking pattern generator \cite{ref1} is used to develop CoM and angular-momentum trajectories. These are integrated into the Whole-body Admittance Control framework \cite{ref2} for simulations in Choreonoid \cite{ref3}.

\section{Conclusions}
\label{char5}  
\renewcommand\arraystretch{2}
This paper presented a novel 3D polygonal semantic mapping approach that significantly enhances the environmental perception of humanoid robots in complex terrains. Our method uses advanced image processing techniques to boost the accuracy, efficiency, and real-time performance of planar surface extraction. Key innovations include the application of anisotropic diffusion filtering to reduce noise and using the RANSAC algorithm for accurate plane fitting, with both processes optimized on the GPU for real-time operation. Our map management system effectively integrates odometry data to construct a detailed global semantic map and incorporates vertical drift compensation to ensure the map's accuracy. Experimental results confirm that our approach surpasses traditional methods in speed, maintaining processing times under $20~ms$ and enhancing map fidelity and usability for robotic navigation. Future work will adapt this mapping system for real-world robot applications, integrating it with real-time motion planning to validate and refine its practical effectiveness in live scenarios.

\bibliographystyle{IEEEtran.bst}
\bibliography{IEEEabrv,bib}
\end{CJK}
\end{document}